\documentclass[twoside]{article}  
\usepackage[T1]{fontenc}
\usepackage{ae,aecompl}
\usepackage[utf8x]{inputenc}
\usepackage{lmodern}
\usepackage[numbers,sort&compress,comma]{natbib}
\usepackage{fullpage}
\usepackage{amsmath}
\usepackage{amssymb}
\usepackage{dsfont}
\usepackage{amsthm}
\usepackage{mathtools}
\usepackage{graphicx}
\usepackage{multirow}
\usepackage{mdwlist}
\usepackage{booktabs}
\usepackage{datetime}

\newif\ifarxiv
\arxivtrue

\def\qed{}

\def\Ds{{\cal D}}
\def\Itm{{\cal I}}
\def\FI{\mathsf{FI}}
\def\TFI{\mathsf{TFI}}
\def\VC{\mathsf{VC}}
\def\EVC{\mathsf{EVC}}

\def\prob{\pi}
\def\tfreq{t_\prob}
\def\range{\mathcal{R}}

\newtheorem{corollary}{Corollary}
\newtheorem{lemma}{Lemma}
\newtheorem{fact}{Fact}
\newtheorem{theorem}{Theorem}

\theoremstyle{definition}
\newtheorem{definition}{Definition}

\begin{document}
\title{Finding the True Frequent Itemsets\thanks{Work supported in part by NSF
grant IIS-1247581. 
\ifarxiv
 This is an extended version of the work that appeared
as~\citep{RiondatoV14}.
\fi 
}
}
\author{Matteo Riondato
\ifarxiv
\thanks{Department of Computer Science, Brown University, Providence, RI, USA.
\url{matteo@cs.brown.edu} . Contact author.}
\fi 
\and Fabio Vandin
\ifarxiv
\thanks{Department of Computer Science, Brown University, Providence, RI, USA
and Department of Mathematics and Computer Science, University of Southern
Denmark, Odense, Denmark. \url{vandinfa@imada.sdu.dk} .}
\fi
}

\date{\today}

\maketitle

\begin{abstract} \small\baselineskip=9pt
   Frequent Itemsets (FIs) mining is a fundamental primitive in data mining. It
   requires to identify 
   all itemsets appearing in at least a fraction 
   $\theta$ of a transactional dataset $\Ds$. Often though, the ultimate goal
   of mining $\Ds$ is not an analysis of the dataset \emph{per se}, but the
   understanding of the underlying process that generated it. 
   Specifically, in many applications $\Ds$ is a collection of samples obtained from an
   unknown probability distribution $\prob$ on transactions, and by extracting
   the FIs in $\Ds$ one attempts to infer itemsets that are
   frequently (i.e., with probability at least $\theta$) generated by $\prob$, which we call the True Frequent Itemsets
   (TFIs). Due to the inherently stochastic nature of the generative process, the
   set of FIs is only a rough approximation of the set of TFIs, as it 
   often contains
   a huge number of \emph{false positives}, 
   i.e., spurious itemsets that are not among the
   TFIs. In this work we design and analyze an algorithm 
   to identify a threshold $\hat{\theta}$ such that the collection of itemsets
   with frequency at least $\hat{\theta}$ in $\Ds$
   contains only TFIs with probability at least $1-\delta$, for
   some user-specified $\delta$. Our method uses 
   results from statistical learning theory involving the (empirical) VC-dimension of the
   problem at hand. This allows us to identify 
   almost all the TFIs without including any false positive. 
   We also experimentally compare our method with the direct mining of $\Ds$ at
   frequency $\theta$ and with
   techniques based on widely-used standard bounds (i.e., the Chernoff bounds)
   of the binomial distribution, and show that our algorithm outperforms these methods and
   achieves even better results than what is guaranteed by the theoretical
   analysis.
 \end{abstract}


{\bf Keywords:} Frequent itemsets, VC-dimension, False positives,
Distribution-free methods, Frequency threshold identification.

\section{Introduction}\label{sec:intro}

The extraction of association rules is one of the fundamental primitives in
data mining and knowledge discovery from large databases~\citep{AgrawalIS93}.
In its most general definition, the problem can be reduced to identifying
frequent sets of items, or \emph{frequent itemsets}, appearing in at
least a fraction  $\theta$ of all transactions in a dataset, where $\theta$ is provided in
input by the user. Frequent itemsets and association rules are not only of
interest for classic data mining applications (e.g., market basket analysis), but
are also useful for further data analysis and mining task, including clustering,
classification, and indexing~\citep{han2006data,HanCXY07}.

In most applications, the set of frequent itemsets is not interesting \emph{per
se}. %
Instead, the mining results 
are used to infer properties of the \emph{underlying process} that generated the
dataset. Consider for example the following scenario: a researcher would like 
to identify frequent associations (i.e., itemsets) between preferences among
Facebook users. To this end, she sets up an online survey 
which is filled out by a \emph{small fraction} of Facebook users (some users may even
take the survey multiple times). Using this information, the researcher wants to
infer the associations (itemsets) that are frequent for the \emph{entire} Facebook
population. In fact, the 
 whole Facebook population and the online survey define the underlying \emph{process} that
generated the dataset \emph{observed} by the researcher. In this work we are
interested in answering the following question: 
how can we use the latter (the observed dataset) to identify itemsets that are
frequent in the former (the whole population)?
This is a very natural question, as is the underlying assumption that the
observed dataset is \emph{representative} of the generating process. For
example, in market basket analysis, 
the observed purchases of customers are used to infer the future
purchase habits 
of all customers while assuming that 
the purchase behavior that generated the dataset is representative of the one
that will be followed in the future.

A natural and general model to describe these concepts 
is to assume that the transactions in the dataset $\Ds$ are \emph{independent
identically distributed} (i.i.d) samples from an \emph{unknown} probability
distribution $\prob$ defined on all possible transactions built on a set of
items. 
Since $\prob$ is fixed, each itemset $A$ has a fixed \emph{probability} $\tfreq(A)$
to appear in a transaction sampled from $\prob$. We call $\tfreq(A)$ the
\emph{true frequency} of $A$ (w.r.t.~$\prob$). The true frequency corresponds
to the fraction of transactions that contain the itemset $A$ among an infinite
set 
of transactions. 
The real goal of the mining process is then to identify itemsets that have
true frequency $\tfreq$ at least $\theta$, i.e., the \emph{True Frequent
Itemsets} (TFIs). 
In the market basket analysis example
, $\Ds$ contains 
the observed purchases of customers, the \emph{unknown} distribution $\prob$
describes the purchase behavior of the customers as a whole, and we want to
analyze $\Ds$ to find the itemsets that have probability (i.e., true frequency) 
at least $\theta$ to be bought by a customer.

Since $\Ds$ represents only a \emph{finite} sample from $\prob$, the set $F$ of frequent itemsets
of $\Ds$ w.r.t.~$\theta$ 
only provides an \emph{approximation} of the True Frequent Itemsets: 
due to the stochastic nature of the generative process 
$F$ 
may contain a number of \emph{false positives}, 
 i.e., itemsets that appear among the frequent itemsets of
 $\Ds$ but whose \emph{true} frequency is smaller than 
$\theta$. 
At the same time, some itemsets with true frequency
greater than $\theta$ may have a frequency in $\Ds$ that is \emph{smaller} than
$\theta$ (\emph{false negatives}), and therefore not be in $F$. This implies
that 
one can not aim at identifying \emph{all and only} the itemsets having true frequency 
at least $\theta$.
Even worse, from the data analyst's point of view, 
there is \emph{no guarantee or bound on the number of false positives} reported
in $F$. 
Consider the following scenario as an example. Let $A$ and $B$ be two (disjoint)
sets of pairs of items. The set $A$ contains 1,000 disjoint pairs, while $B$
contains 10,000 disjoint pairs. Let $\prob$ be such that, for any pair $(a,a')\in
A$, we have $\tfreq((a,a'))=0.1$, and for any pair $(b,b')\in B$, we have
$\tfreq((b,b'))=0.09$. Let $\Ds$ be a dataset of 10,000 transactions sampled from
$\prob$. 
We are interested in finding pairs of items that have true frequency at least 
$\theta=0.095$. 
If we extract the pairs of items with frequency at least $\theta$ in $\Ds$,
it is easy to see that in expectation 
50 of the 1,000 pairs from $A$ will have frequency in $\Ds$ 
\emph{below} $0.095$, and in expectation 
400 pairs from $B$ will have frequency in $\Ds$ 
\emph{above} $0.095$.
Therefore, the set of pairs that have frequency at least $\theta$ in $\Ds$ does
\emph{not} contain 
some of the pairs that have true frequency 
at least $\theta$ 
(false negatives), but 
includes a huge number of 
pairs that have true frequency smaller than $\theta$ (false positives). 

In general, one would like to avoid false positives and 
at the same time find as many TFIs as possible. These are somewhat contrasting
goals, and care must be taken to achieve a good balance between them.
A na\"ive but \emph{overly conservative} method to avoid false positives
involves the use of \emph{Chernoff and union bounds}~\citep{MitzenmacherU05}.
The frequency $f_\Ds(A)$ of an itemset $A$ in $\Ds$ is a random variable with
Binomial distribution $\mathcal{B}(|\Ds|,\tfreq(A))$. It is possible to use
standard methods like the Chernoff and the union bounds to bound the deviation
of the frequencies in the dataset of \emph{all} itemsets from their
expectations. These tools can be used to compute a value $\hat\theta$ such that
the probability that a non-true frequent itemset $B$ has frequency
greater or equal to $\hat\theta$ is at most $1-\delta$, for some
$\delta\in(0,1)$. This method has the following serious drawback: in order to
achieve such guarantee, it is \emph{necessary} to bound the deviation of the
frequencies of \emph{all itemsets possibly appearing in the dataset}~\citep{KirschMAPUV12}. This means
that, if the transactions are built on a set of $n$ items, the union bound must
be taken over all $2^n-1$ potential itemsets, even if some or most of them may
appear with very low frequency or not at all in samples from $\prob$. As a
consequence, the chosen value of $\hat\theta$ is extremely \emph{conservative}, despite
being sufficient to avoid the inclusion of false positives in mining results.  
The collection of itemsets with frequency at least $\hat\theta$ in $\Ds$,
although consisting (probabilistically) only of TFIs, it only contains a
\emph{very small} portion of them, due to the overly conservative choice of
$\hat\theta$. (The results of our experimental evaluation
in Sect.~\ref{sec:experiments} clearly show the limitations of this method.) More
refined algorithms are therefore needed to achieve the correct balance between
the contrasting goals of avoiding false positives and finding as many TFIs as
possible.


\subsection{Our contributions.}
The contributions of this work are the following:
\begin{itemize*}
  \item We formally define the problem of mining the \emph{True Frequent
    Itemsets} w.r.t.~a minimum threshold $\theta$, and we develop and analyze an
    algorithm to \emph{identify a value $\hat{\theta}$ such that, with
    probability at least $1-\delta$, all itemsets
with frequency at least $\hat{\theta}$ in the dataset have true frequency
at least $\theta$}. Our method is completely \emph{distribution-free}, i.e., it
does not make \emph{any} assumption about the unknown generative distribution
$\prob$. 
By contrast, existing methods to assess the significance of frequent patterns after their
extraction 
require a well specified, limited generative model to characterize the
significance of a pattern. When 
additional information about the distribution $\prob$ is available, it can be
incorporated in our method to obtain even higher accuracy.
\item 
We analyse our algorithm using results from \emph{statistical learning theory} and \emph{optimization}. 
We define a range set associated to a collection of itemsets and give an upper
bound to its (empirical) VC-dimension and a procedure to compute this bound,
showing an interesting connection with the Set-Union
Knapsack Problem (SUKP)~\citep{GoldschmidtNY94}. 
To the best of our knowledge, ours is the first work to apply these
techniques to the field of TFIs, and in general the first application of the
sample complexity bound based on \emph{empirical} VC-dimension
to the field of data mining. 
\item We implemented our algorithm and assessed its performances on simulated
  datasets with properties -- number of items, itemsets frequency distribution,
  etc.-- similar to real datasets. We computed the fraction of TFIs contained in the set of frequent itemsets in
  $\Ds$ w.r.t.~$\hat\theta$, and the number of false positives, if any. The
  results show 
  that the algorithm is even \emph{more accurate} than the theory guarantees, since \emph{no
  false positive} 
is reported in any of the many experiments we performed,
  and moreover allows the \emph{extraction of almost all TFIs}. 
  We also
compared the set of itemsets computed by our method to those obtained with the
``Chernoff and union bounds'' method presented in the introduction, and found
that our algorithm \emph{vastly outperforms} it.
\end{itemize*}

\paragraph*{Outline.} 
In Sect.~\ref{sec:prevwork} we review relevant previous contributions.
Sections ~\ref{sec:prelims} and~\ref{sec:range} contain preliminaries to formally define the problem
and key concepts that we will use throughout the work. Our proposed algorithm is
described and analyzed in Sect.~\ref{sec:main}. We present the methodology and
results of our experimental evaluation 
in Sect.~\ref{sec:experiments}. Conclusions and future work can be found 
in Sect.~\ref{sec:concl}. 

\section{Previous work}\label{sec:prevwork}

While the problem of identifying the TFIs has received scant attention in the literature, a number 
of approaches have been proposed to filter the FIs of \emph{spurious patterns}, i.e., patterns that are not
actually \emph{interesting}, according to some interestingness measure. We refer the reader to~\citep[Sect.~3]{HanCXY07}
and~\citep{GengH06} for surveys on different measures.
We remark that, as noted by~\citet{LiuZW11}, that the use of the minimum
support threshold $\theta$, reflecting the level of domain significance, is complementary to the
use of interestingness measures, and that ``statistical significance measures and domain significance
measures should be used together to filter uninteresting rules from different
perspectives''. The algorithm we present can be seen as a method to filter out
patterns that are not interesting according to the measure represented by the
true frequency.

A number of works explored the idea to use statistical properties of the
patterns in order to assess their interestingness. While this is not the focus of our work,
some of the techniques and models proposed are relevant to our framework.
Most of these works are
focused on association rules, but some results can be applied to itemsets. In
these works, the notion of interestingness is related to the deviation between
the observed frequency of a pattern in the dataset and its expected support in a
random dataset generated according to a well-defined probability distribution that can incorporate
prior belief and that can be updated during the mining process to ensure that
the most ``surprising'' patterns are extracted. In many previous works, the
probability distribution was defined by a simple independence model: an item belongs to a
transaction independently from other
items~\citep{SilversteinBM98,MegiddoS98,DuMouchelP01,GionisMMT07,Hamalainen10,KirschMAPUV12}.
In contrast, our work does not impose any
restriction on the probability distribution generating the dataset, with the result that our method is as general as
possible.

\citet{KirschMAPUV12} developed a multi-hypothesis
testing procedure to identify the best support threshold such that the number of
itemsets with at least such support deviates significantly from its expectation
in a random dataset of the same size and with the same frequency distribution
for the individual items. In our work, the minimum threshold $\theta$ is an input
parameter fixed by the user, and we identify a threshold $\hat{\theta}\ge\theta$
to guarantee that the collection of FIs w.r.t.~$\hat{\theta}$ does not contain
any false discovery.

\citet{GionisMMT07} present a method to create random datasets that can act as
samples from a distribution satisfying an assumed generative model. The main
idea is to swap items in a given dataset while keeping the length of the
transactions and the sum over the columns constant. This method is only
applicable if one can actually derive a procedure to perform the swapping in
such a way that the generated datasets are indeed random samples from the assumed
distribution. For the problem we are interested in there such 
procedure is not available. Considering the same generative model,
\citet{Hanhijarvi11} presents a direct adjustment method to bound the
probability of false discoveries by 
taking into consideration the actual number of hypotheses to be tested.

\citet{Webb07} proposes the use of established statistical techniques to
control the probability of false discoveries. 
In one of these methods (called holdout), the available data are split into two parts: one is
used for pattern discovery, while the second is used to verify the significance
of the discovered patterns, testing one statistical hypothesis at a time. A new
method (layered critical values) to choose the critical values when using a
direct adjustment technique to control the probability of false discoveries 
is presented by~\citet{Webb08} and works by exploiting the itemset lattice.
The method we present instead identify a threshold frequency such that all the
itemsets with frequency above the threshold are TFIs. There is no need to test
each itemset separately and no need to split the dataset. 

\citet{LiuZW11} conduct an experimental evaluation of  direct corrections, holdout data,
and random permutations methods to
control the false positives. They test the methods on a very specific problem
(association rules for binary classification). 

In contrast with the methods presented in the works above,
ours does not employ an explicit direct correction depending on the number of
patterns considered as it is  done in traditional multiple hypothesis testing
settings. 
It instead uses the entire available data to obtain more accurate
results,without the need to re-sampling it to generate random datasets or to
split the dataset in two parts, being therefore more efficient computationally.

\section{Preliminaries}\label{sec:prelims}
In this section we introduce the
definitions, lemmas, and tools that we will use throughout the work, providing the details that are
needed in later sections.

\subsection{Itemsets mining.}\label{sec:itemdef}
Given a ground set $\Itm$ of \emph{items}, let $\prob$ be a
probability distribution on $2^{\Itm}$. A \emph{transaction} $\tau\subseteq\Itm$
is a single sample drawn from $\prob$. The \emph{length} $|\tau|$ of
a transaction $\tau$ is the number of items in $\tau$.
A \emph{dataset}
$\Ds$ is a bag of $n$ transactions $\Ds=\{\tau_1,\dots,\tau_n ~:~
\tau_i\subseteq\Itm\}$, i.e., of $n$
\emph{independent identically distributed} (i.i.d.) samples from $\prob$. We
call a subset of $\Itm$ an \emph{itemset}. For any itemset $A$, let
$T(A)=\{\tau\subseteq\Itm ~:~ A\subseteq \tau\}$ be the \emph{support set}
of $A$. 
We define the
\emph{true frequency} $\tfreq(A)$ of $A$ with respect to $\prob$ as the
probability that a transaction sampled from $\prob$ contains $A$:
\[
\tfreq(A) = \sum_{\tau\in T(A)}\prob(\tau)\enspace.
\]


Analogously, given a (observed) dataset $\Ds$, let $T_\Ds(A)$ denote
the set of transactions in $\Ds$ containing $A$. The \emph{frequency} of $A$
in $\Ds$ is the fraction of transactions in $\Ds$ that contain $A$: $f_\Ds(A)=
|T_\Ds(A)|/|\Ds|$. It is easy to see that $f_\Ds(A)$ is the
\emph{empirical average} (and an \emph{unbiased estimator}) for $\tfreq(A)$:
${\mathbf E}[f_\Ds(A)]=\tfreq(A)$.

Traditionally, the interest has been on extracting the set
of \emph{Frequent Itemsets} (FIs) from $\Ds$ with respect to a minimum frequency
threshold $\theta\in(0,1]$~\citep{AgrawalIS93}, that is, the set 
\[
\FI(\Ds,\Itm,\theta)=\{A\subseteq\Itm ~:~ f_\Ds(A)\ge\theta\}\enspace.\]

In most applications the final goal of data mining is to gain a better
understanding of the \emph{process generating the data}, i.e., of the
distribution $\prob$, through the true frequencies $\tfreq$, which are
\emph{unknown} and only approximately reflected in the dataset $\Ds$. Therefore, 
we are interested in finding the itemsets with \emph{true} frequency
$\tfreq$ at least $\theta$ for some $\theta\in(0,1]$. We call these itemsets the
\emph{True Frequent Itemsets} (TFIs) and denote their set as 
\[
\TFI(\prob,\Itm,\theta)=\{A\subseteq\Itm ~:~ \tfreq(A)\ge\theta\}\enspace.\]

If one is only given a \emph{finite} number of random
samples (the dataset $\Ds$) from $\prob$ as it is usually the case, one can not
aim at finding the exact set $\TFI(\prob,\Itm,\theta)$: no assumption can be
made on the set-inclusion relationship between $\TFI(\prob,\Itm,\theta)$ and
$\FI(\Ds,\Itm,\theta)$,
because an itemset $A\in\TFI(\prob,\Itm,\theta)$ may not appear in
$\FI(\Ds,\Itm,\theta)$, and vice versa. One can instead try
to \emph{approximate} the set of TFIs. This is what we are interested in this
work.

\paragraph{Goal.} Given an user-specified
parameter $\delta\in(0,1)$, we aim at providing a threshold
$\hat{\theta}\ge\theta$ such that $\mathcal{C}=\FI(\Ds,\Itm,\hat{\theta})$
\emph{well approximates} $\TFI(\prob,\Itm,\theta)$, in the sense that 
\begin{enumerate*}
  \item With probability at least $1-\delta$, $\mathcal{C}$ does not contain any
    false positive: 
  \[
  \Pr(\exists A\in\mathcal{C} ~:~ \tfreq(A)<\theta)<\delta\enspace.\]
\item $\mathcal{C}$ contains as many TFIs as possible.

  \end{enumerate*}
The 
method we present does not make \emph{any} assumption about 
$\prob$. It uses information from $\Ds$, and guarantees a small probability of
false positives while achieving a high success rate.

\subsection{Vapnik-Chervonenkis dimension.}\label{sec:prelvcdim}
The Vapnik-Chernovenkis (VC) dimension of a collection of subsets of a domain 
is a measure of the complexity or expressiveness of such
collection~\citep{VapnikC71}. 
We outline here some basic definitions and
results and refer the reader to the works of~\citet[Sect.~14.4]{AlonS08} and
\citet[Sect.~3]{BoucheronBL05} for an introduction to VC-dimension and a survey
of recent developments. 

Let $D$ be a domain and $\range$ be a collection of subsets from $D$. We call $\range$ a
\emph{range set on $D$}.
Given $B\subseteq D$, the \emph{projection of $\range$ on $B$} is the set 
$P_\range(B)=\{ B\cap A ~:~ A\in\range\}$. We say that the set $B$ is
\emph{shattered} by $\range$ if $P_\range(B)=2^B$.

\begin{definition}\label{def:empvcdim}
  Given a set $B\subseteq D$, the \emph{empirical Vapnik-Chervonenkis
  (VC) dimension of $\mathcal{A}$ on $B$}, denoted as $\EVC(\range,B)$ is
  the cardinality of the largest subset of $B$ that is shattered by
  $\range$. The \emph{VC-dimension of $\range$} is defined as $\VC(\range)=\EVC(\range,D)$.
\end{definition}


The main application of (empirical) VC-dimension in statistics and learning
theory is in computing the number of samples needed to approximate the
probabilities associated to the ranges through their empirical averages.
Formally, let $X_1^k=(X_1,\dotsc,X_k)$ be a collection of independent
identically distributed random variables taking values in $D$, sampled 
according to some distribution $\nu$ on the elements of $D$.
For a set $A\subseteq D$, let $\nu(A)$ be the probability that a sample from
$\nu$ belongs to the set $A$, and let
\[
\nu_{X_1^k}(A)=\frac{1}{k}\sum_{j=1}^k\mathds{1}_A(X_j),\]
where $\mathds{1}_A$ is the indicator function for the set $A$. The function
$\nu_{X_1^k}(A)$ is the \emph{empirical average} of $\nu(A)$ on $X_1^k$.

\begin{definition}\label{def:eapprox}
  Let $\range$ be a range set on 
  $D$ and $\nu$ be a probability distribution on $D$. For $\varepsilon\in(0,1)$,
  an \emph{$\varepsilon$-approximation to $(\range,\nu)$} is a bag $S$ of
  elements of $D$ such that 
  \[
  \sup_{A\in\range}|\nu(A)-\nu_S(A)|\le\varepsilon\enspace.\]
\end{definition}

An $\varepsilon$-approximation can be constructed by sampling points of
the domain according to the distribution $\nu$, provided an upper bound to the
VC-dimension of $\range$ or to its empirical VC-dimension is known:

\begin{theorem}[Thm.~2.12~\citep{HarPS11}
]\label{thm:eapprox}
  Let $\range$ be a range set on 
  $D$ with $\VC(\range)\le d$, and let $\nu$ be a distribution on $D$. Given
  $\delta\in(0,1)$ and a positive integer $\ell$, let
  \begin{equation}\label{eq:vceapprox}
    \varepsilon = \sqrt{\frac{c}{\ell}\left(d + \log\frac{1}{\delta}\right)}
  \end{equation}
  where $c$ is an universal positive constant. Then, a bag of $\ell$
  elements of $D$ sampled independently according to $\nu$ is an
  $\varepsilon$-approximation to $(\range,\nu)$ with probability at least
  $1-\delta$.
\end{theorem}
\citet{LofflerP09} estimated 
experimentally that the constant $c$ is at most $0.5$.

\begin{theorem}[Sect.~3~\citep{BoucheronBL05}]\label{thm:eapproxempir}
  Let $\range$ be a range set on 
  $D$, and let $\nu$ be a distribution on $D$. Let
  $X_1^\ell=(X_1,\dotsc,X_\ell)$ be a collection of elements from $D$ sampled
  independently according to $\nu$. Let $d$ be an integer such that
  $\EVC(\range,X_1^\ell)\le d$.
  Given $\delta\in(0,1)$, let 
  \begin{equation}\label{eq:evceapprox}
    \varepsilon =
    2\sqrt{\frac{2d\log(\ell+1)}{\ell}}+\sqrt{\frac{2\log\frac{2}{\delta}}{\ell}}.
  \end{equation}
   Then, $X_1^\ell$ is a $\varepsilon$-approximation for $(\range,\nu)$
   with probability at least $1-\delta$.
 \end{theorem}

\section{The range set of a collection of itemsets}\label{sec:range}
In this section we define the concept of a range set associated to a
collection of itemsets and show how to bound the VC-dimension and the
empirical VC-dimension of this range set. We use these definitions and results
to develop our algorithm in later sections.

\begin{definition}\label{def:rangeset}
Given a collection $\mathcal{C}$ of itemsets built on a ground set $\Itm$, the
\emph{range set $\range(\mathcal{C})$ associated to $\mathcal{C}$ is a range
set on $2^\Itm$} containing the support sets of the itemsets in $\mathcal{C}$:
\[\range(\mathcal{C})=\{T(A) ~:~ A\in\mathcal{C}\}\enspace.\]
\end{definition}


\begin{theorem}\label{lem:evcdimupbound}
  Let $\mathcal{C}$ be a collection of itemsets and let $\Ds$ be a dataset. Let
  $d$ be the maximum integer for which there are at least $d$
  transactions $\tau_1,\dotsc,\tau_d\in \Ds$ such that the set
  $\{\tau_1,\dotsc,\tau_d\}$ is an antichain, and each $\tau_i$, $1\le i\le d$,
  contains at least $2^{d-1}$ itemsets from $\mathcal{C}$. 
  Then $\EVC(\range(\mathcal{C}),\Ds)\le d$.
\end{theorem}

\ifarxiv
\begin{proof}
  The antichain requirement guarantees that the set of transactions considered in
  the computation of $d$ could indeed theoretically be shattered. Assume that a
  subset $\mathcal{F}$ of $\Ds$ contains two transactions $\tau'$ and $\tau''$
  such that $\tau'\subseteq\tau''$. Any itemset from $\mathcal{C}$
  appearing in $\tau'$ would also appear in $\tau''$, so there would not be any
  itemset $A\in\mathcal{C}$ such that $\tau''\in T(A)\cap F$ but
  $\tau'\not\in T(A)\cap \mathcal{F}$, which would imply that $\mathcal{F}$ can
  not be shattered. Hence sets that are not antichains should not be
  considered. This has the net effect of potentially resulting in a lower $d$,
  i.e., in a stricter upper bound to $\EVC(\range(\mathcal{C}),\Ds)$.

  Let now $\ell>d$ and consider a set $\mathcal{L}$ of $\ell$ transactions from
  $\Ds$ that is an antichain. Assume that $\mathcal{L}$ is shattered by
  $\range(\mathcal{C})$. Let $\tau$ be a transaction in $\mathcal{L}$.
  The transactions $\tau$ belongs to $2^{\ell-1}$ subsets of $L$. Let
  $\mathcal{K}\subseteq \mathcal{L}$ be one of these subsets. Since
  $\mathcal{L}$ is shattered, there exists an itemset $A\in\mathcal{C}$ such
  that $T(A)\cap \mathcal{L}=\mathcal{K}$. From this and the fact
  that $t\in \mathcal{K}$, we have that $\tau\in T(A)$ or equivalently that
  $A\subseteq\tau$. Given that all the subsets $\mathcal{K}\subseteq\mathcal{L}$
  containing $\tau$ are different, then also all the $T(A)$'s such that
  $T(A)\cap \mathcal{L}=\mathcal{K}$ should be
  different, which in turn implies that all the itemsets
  $A$ should be different and that they should all appear in $\tau$. There are
  $2^{\ell-1}$ subsets $\mathcal{K}$ of $\mathcal{L}$ containing $\tau$,
  therefore $\tau$ must contain at least $2^{\ell-1}$ itemsets from
  $\mathcal{C}$, and this holds for all $\ell$ transactions in $\mathcal{L}$. This is a
  contradiction because $\ell>d$ and $d$ is the
  maximum integer for which there are at least $d$ transactions containing at
  least $2^{d-1}$ itemsets from $\mathcal{C}$. Hence $\mathcal{L}$ cannot be shattered and
  the thesis follows.
  \qed
\end{proof}
\fi

\subsection{Computing the VC-Dimension.}
The na\"ive computation of $d$  according to the definition in Thm.~\ref{lem:evcdimupbound}
requires to scan the transactions one by one, 
compute the number of itemsets from $\mathcal{C}$ appearing in each
transaction, and make sure to consider only itemsets constituting antichains. Given the very large
number of transactions in typical dataset and the fact that the number of
itemsets in a transaction is exponential in its length, this method would be
computationally too expensive. An upper bound to $d$ (and therefore to
$\EVC(\range(\mathcal{C}),\Ds)$) can be computed by solving a
\emph{Set-Union Knapsack Problem} (SUKP)~\citep{GoldschmidtNY94} associated to
$\mathcal{C}$.

\begin{definition}[\citep{GoldschmidtNY94}]\label{def:sukp}
  Let $U=\{a_1,\dotsc,a_\ell\}$ be a set of elements and let
  $\mathcal{S}=\{A_1,\dotsc,A_k\}$ be a set of subsets of $U$, i.e.
  $A_i\subseteq U$ for $1\le i\le k$. Each subset $A_i$, $1\le i\le k$, has an associated
  non-negative \emph{profit} $\rho(A_i)\in\mathbb{R}^+$, and each element $a_j$, $1\le
  j\le\ell$ as an associated non-negative weight $w(a_j)\in\mathbb{R}^+$.
  Given a subset $\mathcal{S}'\subseteq\mathcal{S}$, we define the profit of
  $\mathcal{S}'$ as $P(\mathcal{S}')=\sum_{A_i\in \mathcal{S}'}\rho(A_i)$. Let
  $U_{\mathcal{S}'}=\cup_{A_i\in\mathcal{S}'} A_i$. We
  define the weight of $\mathcal{S}'$ as $W(\mathcal{S}')=\sum_{a_j\in
  U_{\mathcal{S}'}} w(a_j)$. Given a non-negative parameter $c$ that we call
  \emph{capacity}, the \emph{Set-Union Knapsack Problem} (SUKP) requires to find
  the set $\mathcal{S}^*\subseteq\mathcal{S}$ which \emph{maximizes}
  $P(\mathcal{S}')$ over all sets $\mathcal{S}'$ such that $W(\mathcal{S}')\le c$.
\end{definition}

In our case, $U$ is the set of items that appear in the itemsets of
$\mathcal{C}$, $\mathcal{S}=\mathcal{C}$, the profits and the weights are all
unitary, and the capacity constraint is an integer $\ell$. We call this
optimization problem the \emph{SUKP associated to $\mathcal{C}$ with capacity
$\ell$}.
It is easy to see 
that the optimal profit of this SUKP is the maximum number
of itemsets from $\mathcal{C}$ that a transaction of length $\ell$ can contain.  
In order to show how to use this fact to compute an upper bound to
$\EVC(\range(\mathcal{C}),\Ds)$, we need to define some additional terminology. Let
$\ell_1,\dotsc,\ell_w$ be the sequence of the
\emph{transaction lengths} of $\Ds$, i.e., for each value $\ell$
for which there is at least a transaction in $\Ds$ of length $\ell$, there is
one (and only one) index $i$, $1\le i\le w$ such that $\ell_i=\ell$. Assume that
the $\ell_i$'s are labelled in sorted decreasing order:
$\ell_1>\ell_2>\dotsb>\ell_w$. Let now $L_i$, $1\le i\le w$ be the maximum number of
transactions in $\Ds$ that have length at least $\ell_i$ and such that
for no two $\tau'$, $\tau''$ of them we have either $\tau'\subseteq\tau''$ or
$\tau''\subseteq\tau'$. The sequences $(\ell_i)_1^w$ and a sequence $(L_i^*)^w$
of upper bounds to $(L_i)_1^w$ can be computed efficiently with a scan of the
dataset. Let now $q_i$ be the optimal profit of the SUKP associated to
$\mathcal{C}$ with capacity $\ell_i$, and let $b_i=\lfloor \log_2q_i\rfloor +1$.
The following lemma uses these sequences to show how to obtain an upper bound to
the empirical VC-dimension of $\mathcal{C}$ on $\Ds$.

\begin{lemma}\label{lem:sukpevc}
  Let $j$ be the minimum integer for which $b_i\le L_i$. Then
  $\EVC(\mathcal{C},\Ds)\le b_j$. 
\end{lemma}
\ifarxiv
\begin{proof}
  If $b_j\le L_j$, then there are at least $b_j$ transactions which can contain
  $2^{b_j-1}$ itemsets from $\mathcal{C}$ and this is the maximum $b_i$ for
  which it happens, because the sequence $b_1,b_2,\dotsc,b_w$ is sorted in
  decreasing order, given that the sequence $q_1,q_2,\dotsc,q_w$ is. Then $b_j$
  satisfies the conditions of Lemma~\ref{lem:evcdimupbound}. Hence
  $\EVC(\mathcal{C},\Ds)\le b_j$.
  \qed
\end{proof}
\fi

\begin{corollary}\label{lem:sukpvc}
  Let $q$ be profit of the SUKP associated to $\mathcal{C}$ with capacity
  equal to $\ell=|\{a\in\Itm ~:~ \exists A\in\mathcal{C} \mbox{ s.t. } a\in
  A\}|$ ($\ell$ is the number of items such that there is at least one itemset in $\mathcal{C}$ containing
  them).
  Let $b=\lfloor\log_2 q\rfloor + 1$. Then
  $\VC(\range(\mathcal{C}))\le b$. 
\end{corollary}

Solving the SUKP optimally is NP-hard in the general case, although there are known restrictions for
which it can be solved in polynomial time using dynamic
programming~\citep{GoldschmidtNY94}. 
For our case, it is actually \emph{not necessary to compute the optimal
solution} to the SUKP: any upper bound solution for which we can prove that
there is no power of two between that solution and the optimal solution would
result in the \emph{same upper bound} to the (empirical) VC dimension, while
substantially speeding up the computation. This property can be specified in
currently available optimization problem solvers (e.g.~CPLEX), which can then can compute the
bound to the (empirical) VC-dimension very fast even for very large instances
with thousands of items and hundred of thousands of itemsets in $\mathcal{C}$,
making this approach practical. 


The range set associated to $2^\Itm$
is particularly interesting for us. It is possible  
to compute bounds to $\VC(\range(2^\Itm))$ and
$\EVC(\range(2^\Itm), \Ds)$ without having to solve a SUKP.

\begin{theorem}[\citep{RiondatoU12}]\label{thm:empvcdimubfirst}
  Let $\Ds$ be a dataset built on a
  ground set $\Itm$. The
  \emph{d-index} $\mathsf{d}(\Ds)$ of $\Ds$ is the maximum integer $d$ such that
  $\Ds$ contains at least $d$ transactions of length at least $d$ 
  that form an antichain. %
  We have $\EVC(\range(2^\Itm),\Ds)\le \mathsf{d}(\Ds)$.
\end{theorem}



\begin{corollary}\label{thm:vcdimubfirst}
  $\VC(\range(2^\Itm))\le |\Itm|-1$.
\end{corollary}

\citet{RiondatoU12} presented an efficient algorithm to compute an upper bound to
the d-index of a dataset with a single linear scan of the dataset $\Ds$.
The upper bound presented in Thm.~\ref{thm:empvcdimubfirst} is tight:
 there are datasets for which
$\EVC(\range(2^\Itm),\Ds)=\mathsf{d}(\Ds)$~\citep{RiondatoU12}.
This implies that the upper bound presented in Corol.~\ref{thm:vcdimubfirst} is
also tight.


\section{Finding the True Frequent Itemsets}\label{sec:main}
In this section we present an algorithm to identify a threshold $\hat{\theta}$ such
that, with probability at least $\delta$ for some user-specified parameter
$\delta\in(0,1)$, all itemsets with frequency at least $\hat{\theta}$ in $\Ds$
are True Frequent Itemsets with respect to a fixed minimum true frequency
threshold $\theta\in(0,1]$. 
The threshold $\hat{\theta}$ can be used to find a collection
$\mathcal{C}=\FI(\Ds,\Itm,\hat\theta)$ of itemsets such that $\Pr(\exists
A\in\mathcal{C} \mbox{ s.t.~} \tfreq(A)<\theta)<\delta$.
The intuition behind the method is the following. 
Let $\mathcal{B}$ be
the \emph{negative border} of $\TFI(\prob,\Itm,\theta)$, that is the set of itemsets
not in $\TFI(\prob,\Itm,\theta)$ but such that all their proper subsets are in
$\TFI(\prob,\Itm,\theta)$. If we can find an $\varepsilon$ such that $\Ds$ is an
$\varepsilon$-approximation to $(\range(\mathcal{B}),\pi)$ then 
we have that any itemset
$A\in\mathcal{B}$ has a frequency $f_\Ds(A)$ in $\Ds$ less than
$\hat{\theta}=\theta+\varepsilon$, given that it must be $\tfreq(A)<\theta$. By the
antimonotonicity property of the frequency, the same holds for all itemsets that
are supersets of those in $\mathcal{B}$. Hence, the only itemsets that can have
frequency in $\Ds$ greater or equal to $\hat{\theta}=\theta+\varepsilon$ are
those with true frequency at least $\theta$. In the following paragraphs we show
how to compute $\varepsilon$.

Let $\delta_1$ and $\delta_2$ be such that $(1-\delta_1)(1-\delta_2)\ge
1-\delta$. Let $\range(2^\Itm)$ be the range space of all itemsets.
We use Corol.~\ref{thm:vcdimubfirst} (resp.~Thm.~\ref{thm:empvcdimubfirst}) to
compute an upper bound $d'$ to $\VC(\range(2^\Itm))$ (resp.~ $d''$ to
$\EVC(\range(2^\Itm),\Ds)$). Then we can use $d'$ in Thm.~\ref{thm:eapprox} (resp.~$d''$ in
Thm~\ref{thm:eapproxempir}) to compute an $\varepsilon_1'$ (resp.~an
$\varepsilon_1''$) such that $\Ds$ is, with probability at
least $1-\delta_1$, an $\varepsilon_1'$-approximation
(resp.~$\varepsilon_1''$-approximation) to $(\range(2^\Itm),\prob)$.
\begin{fact}
Let
$\varepsilon_1=\min\{\varepsilon_1',\varepsilon_1''\}$. 
With probability at least $1-\delta_1$, $\Ds$ is an
$\varepsilon_1$-approximation to $(\range(2^\Itm),\prob)$.
\end{fact}

We want to find an upper bound the (empirical) VC-dimension of
$\range(\mathcal{B})$. To this end, we use the fact that the negative border of a
collection of itemsets is a \emph{maximal
antichain} on $2^\Itm$. 
Let now $\mathcal{W}$ be the \emph{negative
border} of $\mathcal{C}_1=\FI(\Ds,\Itm,\theta-\varepsilon_1)$, 
$\mathcal{G}=\{A\subseteq\Itm ~:~ \theta-\varepsilon_1\le
f_\Ds(A)<\theta+\varepsilon_1\}$, and $\mathcal{F}=\mathcal{G}\cup\mathcal{W}$.

\begin{lemma}\label{lem:antichains}
  Let $\mathcal{Y}$ be the set of maximal antichains in $\mathcal{F}$. If
  $\Ds$ is an $\varepsilon_1$-approximation to $(\range(2^\Itm),\prob)$, then
  \begin{enumerate*}
    \item
      $\max_{\mathcal{A}\in\mathcal{Y}}\EVC(\range(\mathcal{A}),\Ds)\ge\EVC(\range(\mathcal{B}),\Ds)$,
      and
    \item
      $\max_{\mathcal{A}\in\mathcal{Y}}\VC(\range(\mathcal{A}))\ge\VC(\range(\mathcal{B}))$.
  \end{enumerate*}
\end{lemma}
\ifarxiv
\begin{proof}
  Given 
  that $\Ds$ is an $\varepsilon_1$-approximation to $(\range(2^\Itm),\prob)$, 
  then 
  $\TFI(\prob,\Itm,\theta)\subseteq\mathcal{G}\cup\mathcal{C}_1$. From this and
  the definition of negative border and of $\mathcal{F}$, we have that
  $\mathcal{B})\subseteq\mathcal{F}$. Since $\mathcal{B}$ is a maximal
  antichain, then $\mathcal{B}\in\mathcal{Y}$. Hence the thesis.
  \qed
\end{proof}
\fi

In order to compute upper bounds to $\VC(\range(\mathcal{B}))$ and
$\EVC(\range(\mathcal{B}),\Ds)$ we can solve slightly modified SUKPs 
associated to $\mathcal{F}$ with the additional constraint that the
optimal solution, which is a collection of itemsets, \emph{must be a maximal
antichain}. Lemma~\ref{lem:sukpevc} still holds even for the solutions of these
modified SUKPs. Using these bounds in Thms.~\ref{thm:eapprox}
and~\ref{thm:eapproxempir}, we compute an $\varepsilon_2$ such that, with
probability at least $1-\delta_2$, $\Ds$ is an $\varepsilon_2$-approximation to
$(\range(\mathcal{B}),\prob)$. Let 
\[
\hat{\theta}=\theta+\varepsilon_2\enspace.\]


\begin{theorem}\label{lem:vcfull}
With probability at least $1-\delta$, 
$\FI(\Ds,\Itm,\hat\theta)$ contains no false positives:
\[
\Pr\left(\FI(\Ds,\Itm,\hat\theta)\subseteq\TFI(\prob,\Itm,\theta)\right)\ge 1-\delta\enspace.\]
\end{theorem}
\ifarxiv
\begin{proof}
  Consider the two events $\mathsf{E}_1$=``$\Ds$ is an
  $\varepsilon_1$-approximation for $(\range(2^\Itm),\prob)$'' and
  $\mathsf{E}_2=$``$\Ds$ is an
  $\varepsilon_2$-approximation for $(\range(\mathcal{B}),\prob)$''. From
  the above discussion and the definition of $\delta_1$ and $\delta_2$ it
  follows that the event $\mathsf{E}=\mathsf{E}_1\cap\mathsf{E}_1$ occurs with
  probability at least $1-\delta$. Suppose from now on that indeed $\mathsf{E}$
  occurs.

  Since $\mathsf{E}_1$ occurs, then Lemma~\ref{lem:antichains}
  holds, and the bounds we compute by solving the modified SUKP problems are
  indeed bounds to $\VC(\range(\mathcal{B}))$ and
  $\EVC(\range(\mathcal{B},\Ds))$. 
  Since $\mathsf{E}_2$ also occurs, then for any $A\in\mathcal{B}$ we
  have $|\tfreq(A)-f_\Ds(A)|\le\varepsilon_2$, but given that $\tfreq(A)<\theta$
  because the elements of $\mathcal{B}$ are not TFIs, then we have
  $f_\Ds(A)<\theta+\varepsilon_2$. Because of the antimonotonicity property
  of the frequency and the definition of $\mathcal{B}$, this holds for any
  itemset that is not in $\TFI(\prob,\Itm,\theta)$. Hence, the only itemsets that can have a
  frequency in $\Ds$ at least $\hat{\theta}=\theta+\varepsilon_2$ are the TFIs, so
  $\FI(\Ds,\Itm,\hat{\theta})\subseteq\TFI(\prob,\Itm,\theta)$, which concludes
  our proof.
  \qed
\end{proof}

\paragraph{Exploiting additional knowledge about $\prob$.} 
Our algorithm is completely \emph{distribution-free}, i.e., it does not require
any assumption about the unknown distribution $\prob$. On the other hand, when
information about $\prob$ is available, our method can exploit it to achieve
better performances in terms of running time, practicality, and accuracy. 
For example, in most applications $\prob$ will not generate any transaction
longer than some upper bound $\ell\ll|\Itm|$, and this is know. 
Consider for example an online
marketplace like Amazon: it is extremely unlikely (if not humanly impossible)
that a single customer buys one of each available product. Indeed, given the
hundred of thousands of items on sale, it is safe to assume that all the
transactions will contains at most $\ell$ items, for $\ell\ll|\Itm|$. Other
times, like in an online survey, it is the nature of the process that limits the
number of items in a transaction, in this case the number of questions. A
different kind of information about the generative process may consists in
knowing that some combination of items may never occur, because ``forbidden'' in
some wide sense. Other examples are possible. All these pieces of information
can be used to compute better (i.e., stricter) upper bounds to the VC-dimension
$\VC(\range(2^\Itm))$. For example, if we know that $\prob$ will never generate
transactions with more than $\ell$ items, we can safely say that
$\VC(\range(2^\Itm))\le \ell-1$, a much stricter bound than $|\Itm|-1$ from
Corol.~\ref{thm:vcdimubfirst}. This may result in a smaller $\varepsilon_1$, a smaller 
$\varepsilon$, and a smaller $\hat\theta$, which allows to produce
more TFIs in the output collection. In the experimental evaluation, we show the
positive impact of including additional information may on the performances of
our algorithm.
\fi

\section{Experimental evaluation}\label{sec:experiments}
We conducted an extensive evaluation to assess the performances of the algorithm
we propose. In particular, we used it 
to compute values $\hat\theta$
for a number of frequencies $\theta$ on different datasets, and compared the
collection of FIs w.r.t.~$\hat\theta$ with the collection of TFIs, measuring
the number of false positives 
and the fraction of TFIs that were found.


\begin{table*}[tbp]
  \centering
  \begin{tabular}{llrrr}
\toprule
Dataset & Freq.~$\theta$ & TFIs & Times FPs & Times FNs\\
\midrule
\texttt{accidents} & 0.2 & 889883 & 100\% & 100\% \\
\midrule
\texttt{BMS-POS} & 0.005 & 4240 & 100\% & 100\% \\
\midrule
\texttt{chess} & 0.6 & 254944 & 100\% & 100\% \\
\midrule
\texttt{connect} & 0.85 & 142127 & 100\% & 100\% \\
\midrule
\texttt{kosarak} & 0.015 & 189 & 45\% & 55\% \\
\midrule
\texttt{pumsb*} & 0.45 & 1913 & 5\% & 80\% \\
\midrule
\texttt{retail} & 0.0075 & 277 & 10\% & 20\% \\
\bottomrule
\end{tabular}
\caption{Fractions of times that $\FI(\Ds,\Itm,\theta)$ contained false positives
and missed TFIs (false negatives) over 20 datasets from the same ground truth.}
\label{table:fp}
\end{table*}

\subsection{Implementation.}
We implemented the algorithm in Python 3.3. 
To mine the FIs, we used the C implementation by Grahne
and Zhu~\citep{GrahneZ03}. 
Our solver of choice for the SUKPs was IBM\textsuperscript{\textregistered}
ILOG\textsuperscript{\textregistered} CPLEX\textsuperscript{\textregistered}
Optimization Studio 12.3. 
We run the experiments on a number of machines with x86-64 processors running
GNU/Linux 3.2.0.

\subsection{Datasets generation.}\label{sec:dsgen}
We evaluated the algorithm using pseudo-artificial datasets generated by taking the datasets from the FIMI'04
repository\footnote{\url{http://fimi.ua.ac.be/data/}} as the \emph{ground truth} for the true frequencies
$\tfreq$ of the itemsets. We considered the following datasets:
\texttt{accidents}, 
\texttt{BMS-POS}, \texttt{chess},
\texttt{connect}, \texttt{kosarak}, \texttt{pumsb\textsuperscript{*}}, and
\texttt{retail}. 
These datasets differ in size, number of
items, and, more importantly for our case, distribution of the frequencies of the
itemsets~\citep{GoethalsZ04}. We
created a dataset by \emph{sampling 20 million transactions uniformly at random} from a FIMI
repository dataset. In this way the 
the true frequency of an itemset is its frequency in the
original FIMI dataset. Given that our method to find the TFIs is
distribution-free, this is a valid procedure to establish a ground truth. We used
these enlarged datasets in our experiments, and use the original name of the
datasets in the FIMI repository to annotate the results for the datasets we generated.

\begin{table*}[tbp]
  \centering
  \begin{tabular}{llrcccc}
    \toprule
    & & & \multicolumn{4}{c}{Reported TFIs (Average Fraction)} \\
    \cmidrule(l){4-7}
    & & & \multicolumn{2}{c}{``Vanilla'' (no info)} &
    \multicolumn{2}{c}{Additional Info}\\
    \cmidrule(l){4-5} \cmidrule(l){6-7}
    Dataset & Freq.~$\theta$ & TFIs & CU Method & This Work & CU Method & This Work\\
\midrule
\multirow{8}{*}{\texttt{accidents}} & 0.8 & 149 & 0.838& \bf 0.981
& 0.853& \bf 0.981\\
 & 0.7 & 529 & 0.925& \bf 0.985& 0.935& \bf 0.985\\
 & 0.6 & 2074 & 0.967& \bf 0.992& 0.973& \bf 0.992\\
 & 0.5 & 8057 & 0.946& \bf 0.991& 0.955& \bf 0.991\\
 & 0.45 & 16123 & 0.948& \bf 0.992& 0.955& \bf 0.992\\
 & 0.4 & 32528 & 0.949& 0.991& 0.957& \bf 0.992\\
 & 0.3 & 149545 & & & 0.957& \bf 0.989\\
 & 0.2 & 889883 & &  & 0.957& \bf 0.987\\
\midrule
\multirow{6}{*}{\texttt{BMS-POS}} & 0.05 & 59 & 0.845& \bf 0.938&
0.851& \bf 0.938\\
 & 0.03 & 134 & 0.879& \bf 0.992& 0.895& \bf 0.992\\
 & 0.02 & 308 & 0.847& \bf 0.956& 0.876& \bf 0.956\\
 & 0.01 & 1099 & 0.813& 0.868& 0.833& \bf 0.872\\
 & 0.0075 & 1896 & &  & 0.826& \bf 0.854\\
 & 0.005 & 4240 & &  & 0.762& \bf 0.775\\
\midrule
\multirow{5}{*}{\texttt{chess}} & 0.8 & 8227 & 0.964& \bf 0.991&
0.964& \bf 0.991\\
 & 0.775 & 13264 & 0.957 & \bf 0.990 & 0.957 & \bf 0.990 \\
 & 0.75 & 20993 & 0.957& \bf 0.983& 0.957& \bf 0.983\\
 & 0.65 & 111239 & &  & 0.972& \bf 0.991\\
 & 0.6 & 254944 & &  & 0.970& \bf 0.989\\
\midrule
\multirow{5}{*}{\texttt{connect}} & 0.95 & 2201 & 0.802& \bf 0.951&
0.802& \bf 0.951\\
 & 0.925 & 9015 & 0.881& \bf 0.975& 0.881& \bf 0.975 \\
 & 0.9 & 27127 & 0.893& \bf 0.978& 0.893& \bf 0.978\\
 & 0.875 & 65959 & &  & 0.899& \bf 0.974\\
 & 0.85 & 142127 & &  & 0.918& \bf 0.974\\
\midrule
\multirow{4}{*}{\texttt{kosarak}} & 0.04 & 42 & 0.738& \bf 0.939&
0.809& \bf 0.939\\
 & 0.035 & 50 & 0.720& \bf 0.980& 0.780& \bf 0.980\\
 & 0.025 & 82 & &  & 0.682& \bf0.963\\
 & 0.02 & 121 & &  & 0.650& \bf 0.975\\
 & 0.015 & 189 & &  & 0.641& \bf 0.933\\
\midrule
\multirow{5}{*}{\texttt{pumsb*}} & 0.55 & 305 & 0.791& \bf 0.926&
0.859& \bf 0.926\\
 & 0.5 & 679 & 0.929& \bf 0.998& 0.957& \bf 0.998\\
 & 0.49 & 804 & 0.858& \bf 0.984& 0.907& \bf 0.984\\
 & 0.475 & 1050 & &  & 0.942& \bf 0.996\\
 & 0.45 & 1913 & &  & 0.861& \bf 0.976\\
\midrule
\multirow{5}{*}{\texttt{retail}} & 0.03 & 32 & 0.625& \bf 1.00&
0.906& \bf 1.00\\
 & 0.025 & 38 & 0.842& \bf 0.973& 0.972& \bf 0.973\\
 & 0.0225 & 46 & 0.739& 0.934& 0.869& \bf 0.935\\
 & 0.02 & 55 & &  & 0.882& \bf 0.945\\
 & 0.01 & 159 & &  & 0.902& \bf 0.931\\
 & 0.0075 & 277 & &  & 0.811& \bf 0.843\\
 \bottomrule
 \end{tabular}
  \caption{Recall. Average fraction (over 20 runs) of reported TFIs
  in the output of an algorithm using Chernoff and Union bound and of the one
  presented in this work. For each algorithm we present two versions, one
  (Vanilla) which uses no information about the generative process, and one
  (Add.~Info) in which we assume the knowlegde that the process will not
  generate any transaction longer than twice the size of the longest transaction
  in the original FIMI dataset. In bold, the best result (highest reported
  fraction).}
\label{table:power}
\end{table*}

\subsection{False positives and false negatives in
$\FI(\Ds,\Itm,\theta)$.}
In the first set of experiments we evaluated the performances, in terms of
inclusion of false positives and false negatives in the output, of mining the
dataset at frequency $\theta$. Table~\ref{table:fp} reports  
the fraction of times (over 20 datasets from the same
ground truth) that the set $\FI(\Ds,\Itm,\theta)$ contained false positives
(FP) and was missing TFIs (false negatives (FN)). In most cases, especially when there are
many TFIs, the inclusion of false positives when mining at frequency $\theta$
should be expected. This highlights a need for methods like the one presented in
this work, as there is no guarantee that $\FI(\Ds,\Itm,\theta)$
only contains TFIs. On the other hand,
the fact that some TFIs have frequency in the dataset \emph{smaller} than
$\theta$ (false negatives) points out how one can not aim to extract all and
only the TFIs by using a fixed threshold approach (as the one we present).

\subsection{Control of the false positives (Precision).}\label{sec:fwer}
In this set of experiments we evaluated how well the threshold $\hat\theta$
computed by our algorithm allows to avoid the inclusion of false negatives in
$\FI(\Ds,\Itm,\hat\theta)$. 
To this end, we used a wide range of values for the minimum true frequency
threshold $\theta$ (see Table~\ref{table:power}) and fixed $\delta=0.1$. We
repeated each experiment on 20 different enlarged datasets generated from the
same original FIMI dataset. 
In all the \emph{hundreds} of runs of our algorithms, 
$\FI(\Ds,\Itm,\hat\theta)$ \emph{never} contained \emph{any false
positive}, i.e., \emph{always contained only TFIs}. In other words, the
\emph{precision} of the output was 1.0 in all our experiments. Not 
only our
method can give a frequency threshold to extract only TFIs, but 
it is more
\emph{conservative}, in terms of including false positives, than what the
theoretical analysis guarantees. 

\subsection{Inclusion of TFIs (Recall).}
In addition to avoid false positives in the results, one wants to include as many TFIs
as possible in the output collection. To this end, we assessed what fraction of
the total number of TFIs is reported in $\FI(\Ds,\Itm,\hat\theta)$. Since there
were no false positives, this is corresponds to evaluating the \emph{recall} of
the output collection. We fixed $\delta=0.1$, and considered different values
for the minimum true frequency threshold $\theta$ (see Table~\ref{table:power}).
For each frequency threshold, we repeated the experiment on 20 different
datasets sampled from the same original FIMI dataset, 
and found very small variance in the results.  
We compared the fraction of TFIs that our algorithm included in output with that
included by the ``Chernoff and Union bounds'' (CU) method we presented in
Introduction. We compared two variants of the algorithms: one
(``vanilla'') which makes no assumption on the generative distribution $\prob$,
and another (``additional info'') which assumes that the process will not
generate any transaction longer than twice the longest transaction found in the
original FIMI dataset. Both algorithms can be easily modified to include this
information. In Table~\ref{table:power} we report the average fraction of TFIs
contained in $\FI(\Ds,\Itm,\hat\theta)$. 
We can see that the amount of TFIs
found by our algorithm is always very high: only a \emph{minimal}
fraction (often less than $3\%$) 
of TFIs do not appear in the output. This is explained by the fact that the
value $\varepsilon_2$ computed in our method (see Sect.~\ref{sec:main}) is
always smaller than $10^{-4}$. Moreover our solution
\emph{uniformly outperforms} the CU method, often by a huge margin, since
 our algorithm does not have to take into account all
possible itemsets when computing $\hat\theta$. %
Only partial results are reported for the ``vanilla'' variant because of the
very high number of items in the considered datasets: the mining of the
dataset is performed at frequency threshold $\theta-\varepsilon_1$  and if there
are many items, then the value of $\varepsilon_1$ becomes very high because the
bound to the VC-dimension of $\range(2^\Itm)$ is $|\Itm|-1$, and as a
consequence we have $\theta-\varepsilon_1\le 0$. 
We stress, though, that assuming no knowledge about the distribution $\prob$ is not realistic, and
usually additional information, especially regarding the length of the
transactions, is available and can and should be used. The use of 
additional information gives flexibility to our method and improves its
practicality. Moreover, in some cases, it allows to find an even larger fraction
of the TFIs.

\section{Conclusions}\label{sec:concl}
The usefulness of frequent itemset mining is often hindered by spurious discoveries,
or false positives, in the results. In this work we developed an algorithm to compute
a frequency threshold
$\hat\theta$ such that the collection of FIs at frequency $\hat\theta$ is a
good approximation the collection of True Frequent Itemsets. The threshold is
such that that the probability of reporting \emph{any} false positive
is bounded by a user-specified quantity. We used tools from statistical learning
theory and from optimization to develop and analyze the algorithm. The
experimental evaluation shows that the method we propose
can indeed be used to control the presence of false positives while, at the
same time, extracting a very large fraction of the TFIs from huge datasets. 
There are a number of directions for further research. Among these, we find
particularly interesting and challenging the extension of our method to other
definitions of statistical significance for patterns. 
Also interesting is the derivation of better lower bounds to the VC-dimension
of the range set of a collection of itemsets. Moreover, while this work focuses
on itemsets mining, we believe that it can be extended and generalized to other
settings of multiple hypothesis testing, and give another alternative to existing
approaches for controlling the probability of false discoveries. 



\end{document}